\documentclass[sigconf,nonacm]{acmart}

\setcopyright{none}
\renewcommand\footnotetextcopyrightpermission[1]{}
\usepackage{booktabs}           
\usepackage{threeparttable}     
\usepackage{colortbl}

\AtBeginDocument{%
  }

\begin{document}

\title{GuideEvolve: Self-Evolving Agent Framework\\ with Expert Guidance and 3Ts Mechanisms}


\author{Meihong Jia}
\affiliation{%
  \institution{Amazon}
  \city{Tokyo}
  \country{Japan}}
\email{meihongj@amazon.co.jp}

\author{Yasemin Timar}
\affiliation{%
  \institution{Amazon Web Services}
  \city{Luxembourg}
  \country{Luxembourg}}
\email{yastimar@amazon.com}

\author{Jonathan Taws}
\affiliation{%
  \institution{Amazon Web Services}
  \city{Luxembourg}
  \country{Luxembourg}}
\email{tawsj@amazon.com}





\renewcommand{\shortauthors}{Jia et al.}

\begin{abstract}
This study introduces GuideEvolve, a self-evolving agent framework incorporating expert guidance and three autonomous regulatory mechanisms—collectively termed the 3Ts: \textbf{T}hink Fast and Slow, \textbf{T}hrowback, and \textbf{T}ermination by choice. The framework addresses critical limitations in extant self-evolving agent paradigms through dual innovations: (1) a task-specific expert-evaluation-driven evolutionary pipeline, and (2) the integration of 3Ts mechanisms to facilitate cognitive optimization, performance recovery via temporal rollback, and adaptive termination protocols.

Empirical validation across the MGSM and MMLU benchmarks demonstrates that GuideEvolve attains performance metrics comparable to current state-of-the-art models (89.8\% accuracy versus Gödel Agent's 90.6\% on MGSM; 89.1\% versus 87.9\% on MMLU), while exhibiting notably reduced variance (25–59\%) and enhanced robustness. Furthermore, the framework achieves a 100\% evolution success rate—superior to the baseline’s 82\%—and diminishes required optimization iterations by approximately 50\% (from 30 iterations in baseline to 15.1 and 13.4 across tasks), and provides improved stability for production deployment.


\end{abstract}






\maketitle

\section{Introduction}

Self-evolving agents represent a significant advancement in autonomous AI systems, offering the potential for continuous self-improvement through iterative learning and adaptation. While the Gödel Agent framework \cite{yin2025godel} demonstrated promising results using OpenAI models, its practical implementation faces substantial challenges. This paper addresses these challenges through the lens of Amazon Bedrock and Claude models integration, proposing crucial innovations that enhance the robustness and reliability.

The original implementation using OpenAI models, while groundbreaking, exhibited significant stability issues with 4\% of times experiencing unexpected termination during evolution process and 14\% optimization failure, severely limiting its practical utility in production environments. When replicating the Gödel Agent in Bedrock using Claude models, we identified that the original framework's evaluation feedback mechanism, which passes task questions and solver responses without optimization, frequently leads to immediate token exhaustion during the evolution process. These challenges are further compounded by Bedrock's response formatting requirements, necessitating fundamental architectural adjustments for reliable operation.

To address these challenges, we introduce GuideEvolve agent, featuring two key innovations. First, we introduce a novel guided self-evolution framework, where a specialized LLM-based task expert provides analytical feedback to facilitate the agent's autonomous improvement. The expert module generates task-specific evaluations and recommendations, which are then processed by the self-evolving agent to formulate optimization strategies for the solver. This hierarchical architecture maintains distinct contextual boundaries between evaluation logic and evolution control, enabling efficient token utilization while ensuring focused improvement pathways for task solutions. Notably, rather than implementing direct multi-agent interactions, this framework employs a unidirectional guidance mechanism where the expert's analytical insights are systematically integrated into the agent's self-evolving process, leading to iterative refinements in the solver's operational logic. Second, we develop the 3Ts self-governing mechanisms for the self-evolving agent - Think Fast and Slow, Throwback, and Termination by choice - enabling agents to optimize their cognitive processing, retrieving previous best-performed logic for either recovery from deterioration or facilitate the brainstorm for further improvement on the solutions, and intelligently determine evolution endpoints.

Our framework is designed with practical applications in mind, particularly targeting enterprise use cases such as service improvement recommendations in AWS through Service360 product, and retail business strategy recommendations. By addressing the critical issues of convergence and stability, GuideEvolve agent represents a significant step toward making self-evolving agents viable for practical uses. Through comprehensive resource management and robust error handling, our framework transforms the challenges of Bedrock integration into opportunities for enhanced reliability and performance.

\section{Background}
Self-evolving agents represent an emerging paradigm in artificial intelligence that combines recursive self-improvement with large language models (LLMs). These systems aim to autonomously enhance their problem-solving capabilities through iterative refinement of their solution strategies. The theoretical foundations of self-evolving agents draw from several key areas in AI research. Constitutional AI frameworks \cite{bai2022constitutional} provide principles for maintaining reliable and controlled behavior in autonomous systems. Recent advances in AI planning and reasoning, such as Plan-and-Solve \cite{wang2023plan} and ReAct \cite{yao2023react}, offer insights into structuring decision-making processes. Additionally, developments in memory management and state tracking, exemplified by GraphRAG (\cite{han2024retrieval}) and IRCoT \cite{trivedi2022interleaving}, provide crucial mechanisms for maintaining coherent evolution paths.

In the realm of self-evolving agent architectures, Gödel Agent~\cite{yin2025godel} stands out as a pioneering framework for autonomous self-modification. Its self-referential architecture enables recursive code-level evolution, marking a significant advance in the field of architecturally mutable agent systems. Several related research efforts provide context for Gödel Agent's approach. The Darwin Gödel Machine~\cite{zhang2025darwin} offers a similar paradigm for open-ended evolution, using version control to enable non-linear evolutionary paths. AgentSquare~\cite{shang2025agentsquare} takes a more constrained approach, optimizing modular components within a predefined design space. AlphaEvolve~\cite{novikov2025alphaevolve} applies evolutionary principles to specific algorithmic domains, while ADAS~\cite{hu2024adas} and AFlow~\cite{zhang2024aflow} focus on higher-level agent system design through formal optimization methods.

Among these, Gödel Agent and the Darwin Gödel Machine represent the most comprehensive implementations of recursive self-improvement, which allow for fundamental structural modifications and demonstrate the field's progress towards increasingly adaptable systems. However, current approaches often lack efficient evaluation mechanisms, leading to potential degradation in solution quality over iterations. The Gödel Agent framework, for example, reported stability issues with 4\% of times experiencing unexpected termination and 14\% optimization failure, highlighting the need for more reliable evolution control mechanisms. These challenges are particularly acute in enterprise environments, where system stability and consistent performance are paramount.

\section{GuideEvolve Framework}
\begin{figure*}[htbp]
    \centering
    \includegraphics[width=0.8\linewidth]{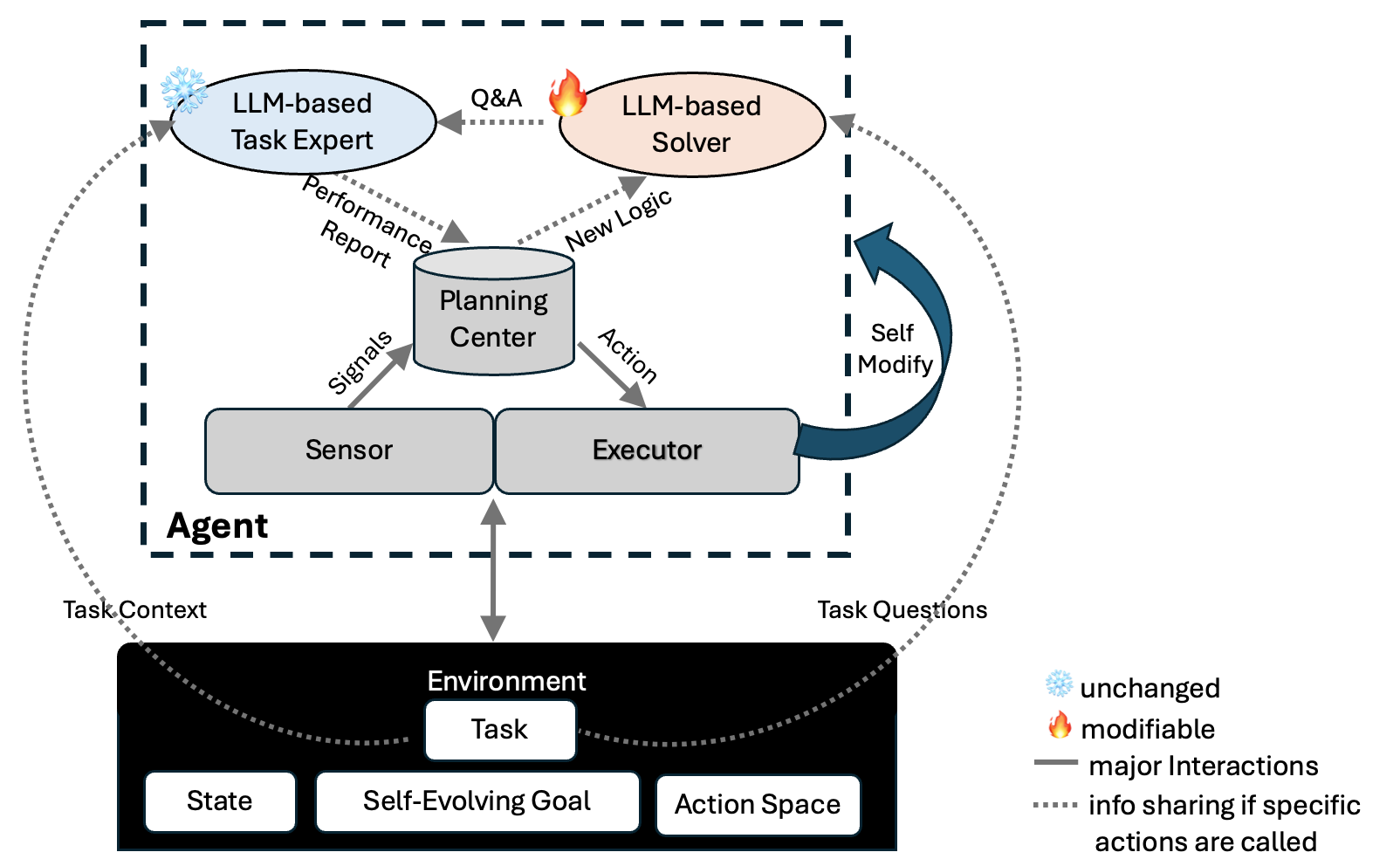}
    \caption{Demonstration of GuideEvolve Agent.}
    \label{fig:diagram_v0}
    \Description{Evolution loop of GuideEvolve Agent}
\end{figure*}

The GuideEvolve framework introduces a self-evolution paradigm that strategically separates task expertise from evolution control, as shown in Figure \ref{fig:diagram_v0}. At its core, a self-evolving agent orchestrates the evolution process by coordinating between two specialized LLM-based modules: a solver module focused on task execution and an expert module dedicated to performance evaluation. This architectural design offers two significant advantages. First, it enables unbiased evaluation through an independent expert system that focuses solely on task-specific assessment without knowledge of the solver's implementation details, while allowing the self-evolving agent to leverage its comprehensive visibility of the environment, historical performance, and solver logic to make informed evolution decisions. Second, this separation addresses the critical challenge of context management in evolution control: by delegating task-specific evaluation to a specialized expert module, the self-evolving agent can more efficiently process and integrate guidance while maintaining essential contextual information about the environment, goals, and evolution strategies. This design fundamentally addresses limitations in existing self-evolving frameworks, where the lack of specialized, task-specific guidance and context overflow often constrain evolution effectiveness.

\textbf{Expert Evaluation Module. } The expert module functions as a specialized analytical consultant, maintaining focused task-specific context derived from dataset descriptions and required capabilities. This module's isolation from the solver's implementation details and evolution state ensures purely objective assessment based on performance outcomes. The expert system processes strategically sampled success and failure cases, analyzing task questions and LLM responses to generate comprehensive performance insights. This stateless architecture serves dual purposes: it maintains evaluation objectivity through isolation from evolution mechanics while significantly reducing the context burden that would otherwise impact the self-evolving agent's efficiency. By processing raw task-solver interactions independently, the expert system can generate focused, high-quality feedback without the overhead of evolution-related contextual information.

\textbf{Guidance Integration and Evolution Mechanism. } The framework implements a structured guidance pipeline that optimizes information flow between its core components. The expert evaluation module generates three critical forms of guidance: quantitative performance metrics, detailed strength/weakness analysis, and actionable insights derived from case studies. The self-evolving agent, as the central control unit, processes this concentrated feedback while maintaining its broader contextual awareness of the environment, historical evolution patterns, and solver behavior. This hierarchical architecture enables efficient information processing: while the evaluation module focuses exclusively on task-specific assessment and recommendations, the self-evolving agent synthesizes these insights with its comprehensive contextual knowledge to formulate targeted evolution strategies. The agent then translates this synthesized understanding into concrete modifications of the solver's operational logic. This modular design not only ensures more focused and effective guidance but also allows the self-evolving agent to manage its context budget more efficiently, dedicating computational resources to critical evolution decisions rather than raw task evaluation processing. The clear separation of evaluation and evolution control enables the agent to maintain a comprehensive view of the evolution trajectory while receiving specialized task-specific guidance.

\section{Understanding Guided Evolution Through an Optimization Lens}

Our intuition for why expert guidance improves both efficiency and robustness is based on an analogy to optimization theory. While the self-evolution process is not 
literally gradient descent, viewing it through this lens helps explain our empirical 
observations and suggests which architectural choices matter most.

We model the evolution as optimizing solver logic $x$ to maximize expected performance 
$f(x)$ on the task distribution:

\begin{equation}
x^* = \operatorname*{arg\,max}_{x} \mathbb{E}[f(x)]
\end{equation}

\noindent\textbf{Key Insight. } The baseline G\"odel Agent approach \cite{yin2025godel} 
resembles \textit{zeroth-order optimization}, where the agent must infer improvement 
directions from noisy function evaluations alone, without explicit gradient information. 
In contrast, GuideEvolve resembles \textit{first-order optimization}, where the expert 
module acts as an oracle providing gradient-like directional guidance. This analogy 
predicts the observed efficiency gains and enhanced stability of our approach.

\subsection{Baseline Evolution as Zeroth-Order Search}

Following the G\"odel Agent framework \cite{yin2025godel}, the self-evolving agent must 
simultaneously evaluate performance and determine update directions. At each iteration 
$t$, the agent's observation of performance is corrupted by context pollution:

\begin{equation}
\tilde{f}(x_t) = f(x_t) + \epsilon_{\text{eval}} + \epsilon_{\text{ctx}}
\end{equation}

where $\epsilon_{\text{eval}}$ represents stochastic sampling noise from evaluating 
on limited training examples, and $\epsilon_{\text{ctx}}$ represents systematic 
observation bias. Concretely, $\epsilon_{\text{ctx}}$ corresponds to the challenge 
of extracting actionable insights when the agent must process raw task-solver 
interactions (e.g., question-answer pairs with their full context) mixed with 
evolution history within a shared context. The agent then computes updates:

\begin{equation}
x_{t+1} = x_t + \alpha \cdot [\text{direction}(\tilde{f}(x_t), \text{history}_t) + \delta_{\text{proc}}^{(\text{high})}]
\end{equation}

where $\delta_{\text{proc}}^{(\text{high})}$ represents processing bias from context 
overflow when maintaining long evolution histories containing raw evaluation data 
alongside decision-making logic. Just as noisy gradient estimates harm optimization 
convergence, the compounding effect of polluted observations ($\epsilon_{\text{ctx}}$) 
and bloated processing context ($\delta_{\text{proc}}^{(\text{high})}$) leads to 
inconsistent evolution trajectories across trials and potential convergence to 
suboptimal solutions.

\subsection{Guided Evolution as First-Order Approximation}

Our framework decomposes the optimization into a two-level structure. The expert 
module acts as an oracle that processes raw performance data in isolation:

\begin{equation}
\begin{aligned}
\tilde{f}_{\text{clean}}(x_t) &= f(x_t) + \epsilon_{\text{eval}} \\
h(x_t) &= \text{expert}(\tilde{f}_{\text{clean}}(x_t), x_t)
\end{aligned}
\end{equation}

The expert processes only task-specific information without access to evolution 
mechanics, producing concentrated guidance $h(x_t)$ analogous to curated gradient 
information. In practice, $h(x_t)$ is a structured report containing 
strengths/weaknesses analysis and actionable recommendations (see Appendix~\ref{sec:expert} 
for an example). This resembles replacing noisy finite-difference gradients with analytical 
gradients in optimization. The self-evolving agent then performs guided updates:

\begin{equation}
x_{t+1} = x_t + \alpha \cdot [\text{direction}(h(x_t), \text{history}_t) + \delta_{\text{proc}}^{(\text{low})}]
\end{equation}

Critically, GuideEvolve eliminates context bias at two levels: (1) \textit{observation 
level}—the agent receives clean guidance $h(x_t)$ rather than polluted observations 
$\tilde{f}(x_t)$ (eliminating $\epsilon_{\text{ctx}}$), and (2) \textit{processing 
level}—the agent's history no longer contains raw evaluation data, substantially 
reducing context load such that $\delta_{\text{proc}}^{(\text{low})} \ll 
\delta_{\text{proc}}^{(\text{high})}$.

\subsection{Predicted Benefits from the Analogy}

This optimization perspective generates three testable predictions:

\textbf{Improved Accuracy through Cleaner Signals. } The expert's context 
isolation eliminates systematic observation bias, providing cleaner directional 
signals. For knowledge-intensive tasks (MMLU), this enables discovery of better local 
optima through more accurate gradient approximations, resulting in superior convergence:

\begin{equation}
\mathbb{E}[f(x^*_{\text{GuideEvolve}})] \geq \mathbb{E}[f(x^*_{\text{baseline}})]
\end{equation}

For tasks with fundamental capability constraints (e.g., MGSM requiring symbolic 
computation), both approaches converge to similar accuracy plateaus determined by 
underlying model capabilities.




\textbf{Enhanced Stability through Variance Reduction. } The dual-level 
elimination of context bias should fundamentally improve trajectory consistency across 
independent trials. We can conceptualize the variance in final performance as composed of multiple sources:

\begin{equation}
\begin{aligned}
\text{Var}[f(x^*)] \approx & \underbrace{\sigma^2_{\text{eval}}}_{\text{from } \epsilon_{\text{eval}}} + 
\underbrace{\sigma^2_{\text{ctx}}}_{\text{from } \epsilon_{\text{ctx}}} \\
& + \underbrace{\sigma^2_{\text{proc}}}_{\text{from } \delta_{\text{proc}}} + 
\underbrace{\sigma^2_{\text{path}}}_{\text{trajectory randomness}}
\end{aligned}
\end{equation}

Under this conceptual model, GuideEvolve should reduce three of the four terms: (a) $\sigma^2_{\text{ctx}}$ 
eliminated through expert's clean observation (removing $\epsilon_{\text{ctx}}$), 
(b) $\sigma^2_{\text{proc}}$ substantially reduced via lean evolution history without 
raw evaluation data (reducing $\delta_{\text{proc}}^{(\text{high})}$ to 
$\delta_{\text{proc}}^{(\text{low})}$), and (c) $\sigma^2_{\text{path}}$ reduced 
via consistent, unbiased guidance signals. Note that $\sigma^2_{\text{eval}}$ 
should remain unchanged as both approaches use identical sampling strategies.

\textbf{Faster Convergence through Informed Search. } The first-order-like 
guidance $h(x_t)$ should enable more efficient exploration compared to zeroth-order 
search with polluted observations and bloated context:

\begin{equation}
\text{Iter.}_{\text{GuideEvolve}} \ll \text{Iter.}_{\text{baseline}}
\end{equation}

\subsection{Empirical Support and Scope of the Analogy}


Our experimental results (Section~\ref{sec:exp}) are consistent with these predictions. 
The accuracy patterns across tasks match the expectation that cleaner signals benefit 
knowledge-intensive tasks (MMLU) more than capability-constrained tasks (MGSM). The 
observed 25-59\% variance reduction (Table~\ref{tab:acc}) is consistent with the predicted 
elimination of $\sigma^2_{\text{ctx}}$ and reduction of $\sigma^2_{\text{proc}}$ and 
$\sigma^2_{\text{path}}$, while $\sigma^2_{\text{eval}}$ remains present as both 
approaches use identical sampling strategies. The 50\% reduction in required iterations 
(Table~\ref{tab:rob}) supports the informed search prediction. This optimization lens—while 
not a formal proof—provides valuable intuition for why specific architectural choices 
(expert isolation, lean context management, and clean information flow) translate into 
the observed stability and efficiency improvements.

\section{Self-Governing Evolution Mechanisms (3Ts)} 

Our framework introduces three critical self-governing mechanisms - Think fast and slow, Throwback, and Termination by Choice (3Ts) - that enable autonomous and efficient evolution control.

\textbf{Think, Fast and Slow. } Our framework implements a dual-mode cognitive processing paradigm through prompt engineering. "Fast thinking" enables efficient execution of multiple low-risk actions in sequence, while "slow thinking" enforces careful step-by-step planning for resource-intensive or error-prone operations. This balance optimizes resource utilization while maintaining solution quality through dynamic planning strategies. Notably, performance evaluation through the expert module is only initiated after confirmed successful modifications to the solver's logic, representing a key "slow thinking" checkpoint in the evolution process. The details can be found in goal prompt in Appendix \ref{sec:goal}.

\textbf{Throwback. } The throwback mechanism represents a crucial innovation in evolution control, extending beyond simple temporal rollback functionality. The system maintains two distinct tracking mechanisms: comprehensive performance monitoring across the entire evolution history, and state preservation of the solver logic at its previous best-performing state. While the performance tracking provides a complete evolutionary trajectory, the selective state preservation minimizes storage overhead while maintaining the capability for both defensive rollback and proactive learning of the self-evolving agent. This design choice reflects our focus on efficient memory utilization while ensuring the agent can either revert to proven optimal states or analyze previous success factors for continued improvement.

\begin{figure*}[htbp]
    \centering
    \includegraphics[width=\linewidth]{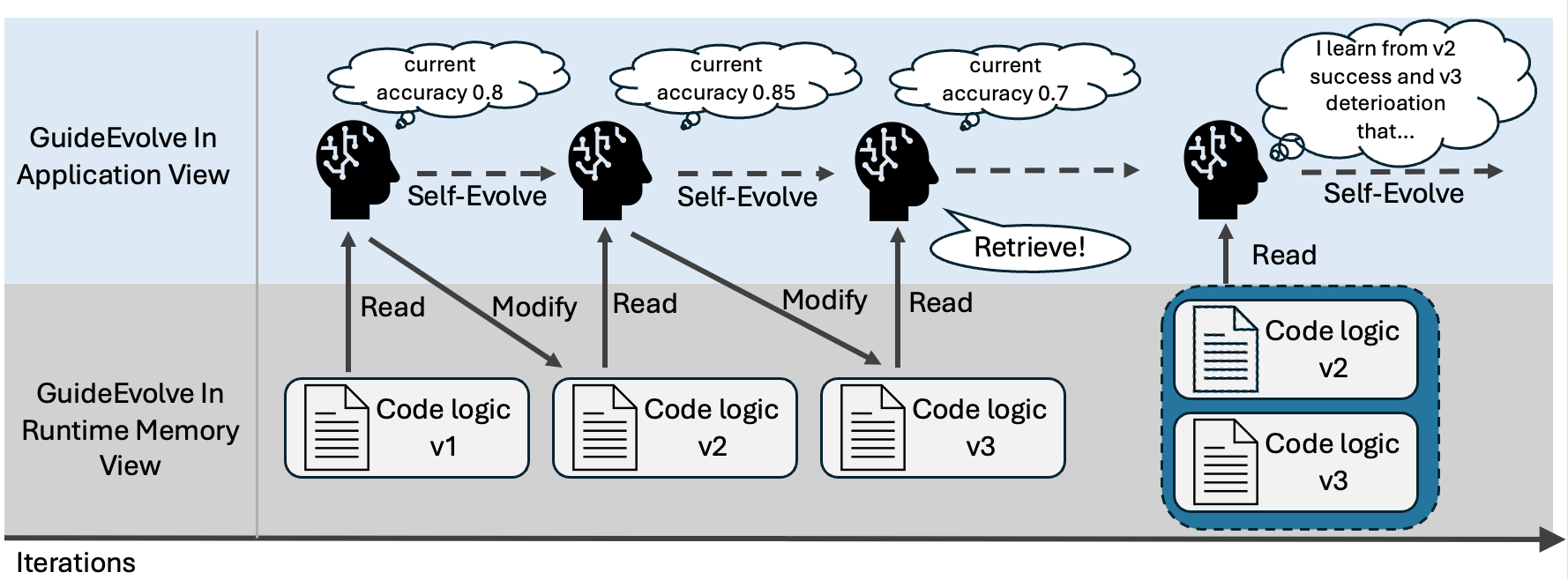}
    \caption{An illustration of implementation of GuideEvolve that calls the throwback action.}
    \Description{An illustration of implementation of GuideEvolve that calls the throwback action.}
    \label{fig:retrieve}
\end{figure*}

We observe three distinct patterns in the agent's autonomous throwback decisions: 1) Strategic state examination, where the agent revisits previous high-performing states to identify untapped optimization opportunities; 2) Defensive rollback in response to significant performance deterioration, restoring the solver to a known optimal state; 3) Context-driven recall during progressive message reduction, where the agent strategically retrieves and analyzes historical solver logic to maintain evolution coherence. An example of throwback is illustrated in Figure \ref{fig:retrieve}.

This mechanism differs fundamentally from traditional tree-based approaches by enabling flexible state examination without enforcing strict hierarchical relationships between states. The agent can autonomously decide when and how to utilize historical information, whether for performance recovery or optimization insight.

\textbf{Termination By Choice. } The framework implements a flexible termination mechanism that balances autonomous decision-making with performance guarantees, as illustrated in Appendix \ref{sec:terminate}. Our observations reveal emerging behavioral patterns in self-evolving agents: some "lazy" agents that attempt early termination after minimal trials, and some "diligent" agents that persist well beyond suggested termination conditions. This demonstrates the full autonomy of termination by agent’s choice. To prevent premature termination, especially from "lazy" agents, the system implements validation checks that provide feedback such as "you need to have at least five trials, keep going" to enforce further exploration of the solution space. 


\section{Resource Management System}
The integration in Bedrock using Claude models necessitates careful management of token limitations and response completeness. Our framework implements a comprehensive resource management system that transforms these constraints into opportunities for optimization.

\textbf{Token Budget Optimization. } Our system employs a dynamic token allocation strategy that prioritizes essential operations while maximizing efficiency. The framework reserves token budget primarily for tool calls — the critical mechanism for action execution — while maintaining system messages as immutable components. A dynamic sliding window manages action history, automatically adjusting output token limits based on input context size. This preventative approach optimizes token utilization while ensuring tool call generation under varying context loads.

\textbf{Response Validation and Context Management. } The framework implements a two-tier approach to handle potential token overflow scenarios. First, it leverages Bedrock's structured response format to implement robust completion detection through JSON validation and tool call verification. When token limitations lead to response truncation, the system activates a progressive context reduction mechanism that systematically preserves essential information while reducing context size. This context reduction process operates sequentially: 1) Removal of older action history messages while maintaining recent; 2) Progressive increase in reduction scope if truncation in response persists; 3) Maintenance of minimal viable context with at least one recent action message. 

This integrated approach, coupled with the framework's throwback capability, ensures operational continuity while minimizing information loss. The system effectively balances immediate operational needs with long-term evolution stability, transforming platform constraints into mechanisms for enhanced reliability.

\section{Experiments Results} \label{sec:exp}
We evaluate GuideEvolve on two benchmark datasets: Mathematics Grade School Multi-source (MGSM) and Massive Multitask Language Understanding (MMLU). MGSM tests multi-step mathematical reasoning and cross-lingual problem-solving capabilities, while MMLU evaluates comprehensive knowledge and problem-solving abilities across 57 diverse academic and professional domains through multiple-choice questions.


\textbf{Experimental Setup. } We conduct our experiments using Anthropic's Claude 3.5 Sonnet v2 as the default initial model for GuideEvolve's components (self-evolving agent, solver, and expert module). The self-evolving agent can autonomously decide to switch models among Anthropic Claude 3.5 Haiku, Anthropic Claude 3.5 Sonnet v2 (implementation identifier: \verb|SONNET_V36|), and Anthropic Claude 3.7 Sonnet (\verb|SONNET_V37|) during the evolution process. Note that the model selected plays a critical role in performance, as also reported in the original G\"odel agent paper (\cite{yin2025godel}).

Following G\"odel's experimental protocol for fair comparison, we evaluate on 800 test samples and perform 6 independent trials for each task, where each trial represents a complete self-improvement process with 20 training samples for validation during evolution. GuideEvolve autonomously determines when to terminate based on either reaching a performance plateau or a maximum of 30 iterations, though in practice, it never requires the full 30 iterations. All quantitative performance results reported in Tables \ref{tab:acc} and \ref{tab:rob} in Section \ref{sec:acc} are based on this full experimental setup.


\textbf{Note on Experimental Scales. } We use two experimental configurations in this paper, summarized in Table~\ref{tab:exp_config}. Configuration A (full-scale) generates all quantitative benchmark results. Configuration B (reduced-scale) enables efficient visualization and ablation analysis without overwhelming complexity. The higher accuracies in Configuration B visualizations (e.g., up to 0.97) reflect the smaller test scale and should not be compared with Configuration A benchmarks.

\begin{table}[h]
\centering
\caption{Experimental Configurations}
\begin{tabular}{@{}lccl@{}}
\toprule
\textbf{Config} & \textbf{Train} & \textbf{Test} & \textbf{Used In} \\
\midrule
A: Full-scale & 20 & 800 & Tables~\ref{tab:acc}, \ref{tab:rob} \\
B: Reduced-scale & 5 & 100 & Figs~\ref{fig:case}, \ref{fig:action_cnt}, \ref{fig:acc_prog}; Table~\ref{tab:ablation} \\
\bottomrule
\end{tabular}
\label{tab:exp_config}
\end{table}

\textbf{Baseline Approaches. } We compare GuideEvolve against both traditional hand-designed agent systems and advanced learning-based approaches\footnote{The results of baseline models refer to Gödel Agent Paper(\cite{yin2025godel}).}. The hand-designed methods include Chain-of-Thought (CoT) (\cite{NEURIPS2022_9d560961}), which implements step-by-step reasoning, and its extension CoT with Self-Consistency (\cite{wang2022self}), which generates multiple solution paths to select the most consistent answer. Self-Refine (\cite{madaan2023self}) enables iterative self-correction, while LLM-Debate (\cite{du2023improving}) facilitates multi-agent debate for diverse perspectives. Step-back Abstraction (\cite{zheng2023take}) emphasizes understanding fundamental principles before detailed problem-solving, Quality-Diversity (\cite{lu2024ai}) combines various solution approaches, and Role Assignment (\cite{xu2023expertprompting}) leverages different perspectives through specific LLM roles. Given the limitations of these fixed algorithms in handling dynamic scenarios, we also compare against Meta Agent Search (\cite{hu2024automated}), the state-of-the-art in automated agent design. Additionally, we benchmark against Gödel (\cite{yin2025godel}), the autonomous self-evolving framework that inspired GuideEvolve, using both its GPT-3.5 (base) and GPT-4.0 (free) implementations. In addition, for a fair comparison without LLM model differences to understand the impact of task expert and 3Ts, we also carried out an ablation study in Appendix \ref{sec:ablation}. 

\subsection{Performance Analysis} \label{sec:acc}

\begin{table} 
\centering
\begin{threeparttable}
    \caption{Accuracy (\%) Comparison of Agent Systems on MGSM and MMLU tasks}
    \begin{tabular}{@{}lrr@{}}
    \toprule
        \textbf{Agent} & \textbf{MGSM} & \textbf{MMLU} \\
    \midrule
        \multicolumn{3}{c}{\cellcolor{gray!20}\textbf{Hand-Designed Agent Systems}} \\
        Chain-of-Thought\cite{NEURIPS2022_9d560961}  & 28.0 ± 3.1 & 65.4 ± .3 \\
        COT-SC\cite{wang2022self}  & 28.2 ± 3.1 & 65.9 ± 3.2 \\
        Self-Refine\cite{madaan2023self}  & 27.5 ± 3.1 & 63.5 ± 3.4 \\
        LLM Debate\cite{du2023improving}  & 39.0 ± 3.4 & 65.6 ± 3.3 \\
        Step-back-Abs\cite{zheng2023take}  & 31.1 ± 3.2 & 65.1 ± 3.3 \\
        Quality-Diversity\cite{lu2024ai} & 23.8 ± 3.0 & 65.1 ± 3.3 \\
        Role Assignment\cite{xu2023expertprompting}  & 30.1 ± 3.2 & 64.5 ± 3.3 \\
        \midrule
        \multicolumn{3}{c}{\cellcolor{gray!20}\textbf{Meta-Learning Optimized Agents}} \\
        Meta Agent Search\cite{hu2024automated} & 53.4 ± 3.5 & 69.6 ± 3.2 \\
    \midrule
        \multicolumn{3}{c}{\cellcolor{gray!20}\textbf{Autonomous Self-evolving Agents}} \\
        Gödel-base-GPT3.5\cite{yin2025godel}\tnote{a} & 64.2 ± 3.4 & 70.9 ± 3.1 \\
        Gödel-free-GPT4.0\cite{yin2025godel}\tnote{b} & \textbf{90.6 ± 2.0} & 87.9 ± 2.2 \\
        GuideEvolve-Claude (ours) & 89.8 ± 1.5 & \textbf{89.1 ± 0.9}\\
    \bottomrule
    \end{tabular}
    \begin{tablenotes}
    \item[a] "base": G\"odel agent with GPT-3.5-turbo as base model
    \item[b] "free": G\"odel agent with unrestricted model selection (GPT-4.0)
    \end{tablenotes}
    \label{tab:acc}
\end{threeparttable}
\end{table}

\begin{table}
\centering
\begin{threeparttable}
\caption{Comparison of Robustness and Evolution Iterations Required}
\begin{tabular}{@{}lrr@{}}
\toprule
\textbf{Agent} & \textbf{Success rate} \tnote{*} & \textbf{Iter.} \\
\midrule
\multicolumn{3}{c}{\cellcolor{gray!20}\textbf{MGSM task}} \\
Gödel agent\cite{yin2025godel} & 82\% & 30.0 \\
GuideEvolve (ours) & \textbf{100\%} & \textbf{15.1} \\
\midrule
\multicolumn{3}{c}{\cellcolor{gray!20}\textbf{MMLU task}} \\
Gödel agent\cite{yin2025godel} & unreported & 30.0 \\
GuideEvolve (ours) & \textbf{100\%} & \textbf{13.4} \\
\bottomrule
\end{tabular}
\begin{tablenotes}
\item[*] Success rate refers to autonomous termination of evolution and a better solution achieved than the initial state.
\end{tablenotes}
\label{tab:rob}
\end{threeparttable}
\end{table}


\textbf{Improved Stability with Competitive Accuracy. } As shown in Table \ref{tab:acc}, we report the test accuracy along with the 95\% bootstrap confidence intervals on test sets, where bold indicates the best performing method. GuideEvolve demonstrates improved stability across both tasks while maintaining competitive accuracy. Across 6 independent trials with 800 test samples per task (identical to G\"odel's experimental setup), GuideEvolve achieves 89.8\% accuracy ($\pm$1.5\%) on MGSM, delivering performance comparable to G\"odel-free (90.6\% $\pm$2.0\%) but with \textbf{25\% lower variance}. For MMLU, GuideEvolve reaches 89.1\% accuracy ($\pm$0.9\%), surpassing G\"odel-free's 87.9\% ($\pm$2.2\%) by 1.2 percentage points while achieving \textbf{59\% lower variance}.

The performance differences across tasks reveal important insights into domain-specific characteristics of self-evolution. For mathematical reasoning tasks (MGSM), while the self-evolving agent successfully explores solver prompts and workflow modifications, the inherent limitations of LLMs in mathematical computation without external tools constrain the achievable performance gains, resulting in comparable accuracy to the baseline. In contrast, for knowledge-intensive verbal reasoning tasks (MMLU), the self-evolution process demonstrates greater efficacy in optimizing solver logic through prompt engineering and reasoning strategies, leading to measurable accuracy improvements over the baseline. These results align with the intuition that self-evolution mechanisms are more effective in domains where linguistic and reasoning optimizations can substantially impact performance, whereas domains requiring symbolic computation or external tool integration present fundamental capability boundaries.


The observed accuracy and variance improvements are primarily attributable to our architectural innovations rather than underlying model differences. Our ablation study (Table~\ref{tab:ablation}) confirms that removing the expert module and throwback mechanism—while maintaining the same Claude models—results in 17\% accuracy decrease and substantial performance degradation, isolating the framework's contribution. These improvements stem from two key mechanisms: First, our task expert module provides objective performance evaluation and actionable proposals, efficiently guiding the self-evolving agent towards better performance, in contrast to G\"odel's approach of relying on the agent to self-evaluate raw task-solver interactions, which often leads to context overflow and potentially biased judgments. Second, GuideEvolve's throwback mechanism enables the agent to revisit and analyze its previous successful strategies, facilitating comparative learning from its own successes and failures. This self-introspection capability becomes particularly valuable when the agent encounters performance plateaus, allowing it to draw insights from its evolution history to push performance boundaries further.


\textbf{Enhanced Robustness. } In Table \ref{tab:rob}, GuideEvolve demonstrates robustness with a 100\% success rate in evolution convergence and finding an improved solution over initial state for both tasks. This perfect convergence rate significantly outperforms G\"odel's 82\% success rate on MGSM (with unreported convergence on MMLU), where G\"odel experiences stability issues in 4\% of trials (unexpected termination during evolution) and optimization failures in 14\% of trials. Our method's success rate across both tasks underscores the stability and reliability of our guided evolution approach. This robustness is achieved through two features: the throwback functionality enables the agent to recover from suboptimal evolution paths by rolling back to previous successful states which prevents optimization failures, while termination by choice allows the agent to intelligently halt the evolution process when reaching a performance plateau, avoiding deterioration from forced continuation beyond the agent's current capability limits.

\textbf{Higher Evolution Efficiency. } GuideEvolve demonstrates remarkable efficiency in its self-improvement process, as shown in Table~\ref{tab:rob}. It requires only 15.1 iterations for MGSM and 13.4 iterations for MMLU on average, compared to G\"odel's 30 iterations—a reduction of approximately 50\% and 55\% in evolving effort respectively. This efficiency is particularly noteworthy given that GuideEvolve achieves competitive or superior accuracy in fewer iterations. This efficiency boost is attributed to: 1) expert guidance facilitates faster convergence towards the best solver; 2) GuideEvolve's intelligent self-regulation capability to autonomously determine termination based on performance plateaus avoids unnecessary iterations.

\subsection{Evolution Traces and Discussions} \label{sec:trace}
As noted in the experimental setup, the following analyses use the reduced-scale visualization setup (5 training, 100 test samples) to illustrate mechanisms clearly.

\begin{figure}[htbp]
    \centering
    \includegraphics[width=\columnwidth]{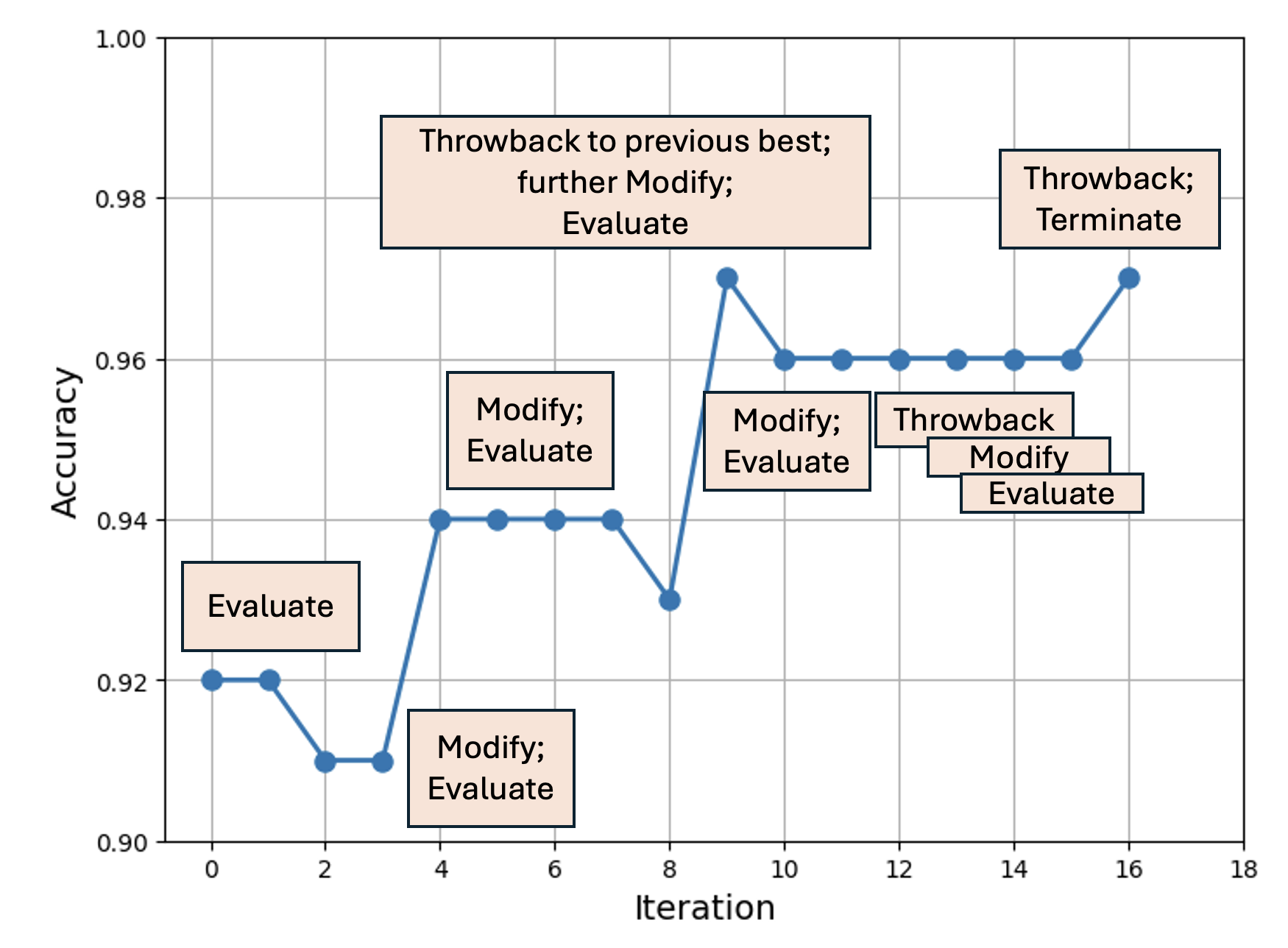}
    \caption{One example evolution of GuideEvolve Agent on MGSM.}
    \label{fig:case}
    \Description{One example evolution of GuideEvolve Agent on MGSM.}
\end{figure}

\textbf{Case Study. } To illustrate GuideEvolve's operation, we examine a representative evolution instance on the MGSM task. Visualization of this instance's evolution path can be found in Figure \ref{fig:case}. Starting with an initial solver accuracy of 0.9 on the test set, the agent first perceives its environment to understand its current state. In its first evolution step, the agent modifies the solver to generate more responses and implement a verification mechanism to select the most reliable answer. Despite achieving 100\% accuracy on the validation set, the test set accuracy initially drops to 0.91, revealing a challenging overfitting scenario. The agent continues evolution despite reaching validation set perfection, avoiding early termination through our minimum iteration requirement. This constraint proves valuable as subsequent iterations explore different strategies, such as implementing few-shot learning, which improves test accuracy to 0.94. Through iterations 4-8, the agent experiments with various modifications, revealing two distinct scenarios: when validation failures occur, the task expert provides highly effective and targeted guidance; however, in overfitting cases (100\% validation accuracy but low test performance), the task expert's effectiveness is limited due to lack of observable failures, leading to less efficient exploration by the self-evolving agent. A crucial breakthrough occurs at iteration 9 when the agent calls throwback and modifies previous best-performing logic, achieving 0.97 accuracy. After further exploration without additional improvements, the agent autonomously terminates at iteration 18, exemplifying the framework's ability to balance exploration with efficient termination.

\begin{figure}[htbp]
    \centering
    \includegraphics[width=\columnwidth]{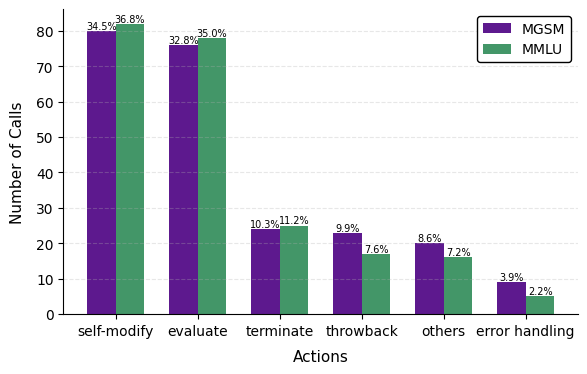}
    \caption{GuideEvolve: Frequency of actions taken by task}
    \label{fig:action_cnt}
    \Description{GuideEvolve: Frequency of actions taken by task presented}
\end{figure}

\textbf{Action Analysis. } Across all trials of GuideEvolve on both tasks, the number of actions taken by the agent reveals consistent behavioral tendencies across tasks, as shown in Figure \ref{fig:action_cnt}. The agent predominantly engages in self-modification and evaluation actions, which directly align with its core objective of solver improvement. Self-modification allows the agent to refine its problem-solving logic, while evaluation leverages task expert guidance to assess these modifications' effectiveness. Notably, the two introduced mechanisms - throwback and self-termination - rank as the next most frequent actions, indicating their significant role in the agent's evolution strategy. This action distribution demonstrates that the agent actively utilizes these auxiliary mechanisms to navigate its self-evolution path more effectively, supporting the value of incorporating these additional control mechanisms in the agent's decision-making repertoire.

\textbf{Accuracy Progression Analysis. } Figure \ref{fig:acc_prog} in Appendix \ref{sec:case} demonstrates distinct paths of self-improvement for three independent GuideEvolve agents, all initiating from a notably high baseline accuracy of 0.9 on the MGSM task. Despite the challenging nature of improving upon such a strong starting point, each agent successfully enhanced its problem-solving capabilities through different trajectories. The first agent (orange line) exhibited rapid convergence, reaching an accuracy of 0.96 within just 8 iterations before self-terminating. The second agent (green line) showed steady improvement over 18 iterations, ultimately achieving an accuracy of 0.97. The third agent (blue line) demonstrated resilience by recovering from a temporary complete performance drop (accuracy of 0) - likely due to output formatting issues of the solver - and continued its evolution to reach a peak accuracy of 0.97 after 19 iterations. These diverse evolution patterns highlight the agent's ability to autonomously navigate and recover from setbacks while consistently finding paths to improved performance.

\subsection{Ablation Study} \label{sec:ablation}

\begin{table}
\centering
  \begin{threeparttable}
    \caption{Ablation Study Results}
    \small
    \begin{tabular}{@{}lrrr@{}}
    \toprule
    \textbf{Ablation} & \textbf{Success (\%)} & \textbf{Accuracy (\%)} & \textbf{Iterations} \\
    \midrule
    w/o expert nor throwback & 80 {\scriptsize ↓20} & 79 {\scriptsize ↓17} & 22.5 {\scriptsize ↑8.9} \\
    \bottomrule
    \end{tabular}
    \label{tab:ablation}
  \end{threeparttable}
\end{table}
To isolate the impact of the task expert module and throwback mechanism, we conduct an ablation study using the reduced-scale setup (5 training, 100 test samples) described in the note of Section \ref{sec:exp}. This smaller scale enables efficient comparison of architectural variations while maintaining representative performance patterns.

Table \ref{tab:ablation} demonstrates the critical role of task expert guidance and throwback mechanisms in our system. We chose a baseline that maintains all other features while removing these two components, as this represents the minimal viable configuration required for evolution with bedrock and Claude models. The results show significant performance degradation across all metrics: success rate drops by 20\%, accuracy decreases by 17\%, and the number of iterations increases by 8.9. Without task expert guidance, the agent struggles with self-evaluation of solutions, while the absence of throwback functionality makes it difficult to recover from performance deterioration. These limitations, combined with frequent context overflow issues, can lead to fatal errors that prematurely terminate the evolution process, highlighting the importance of these mechanisms for robust and efficient self-evolution.

\subsection{Limitations and Threats to Validity}

\textbf{Model Confound. } Our primary comparison evaluates Claude-based GuideEvolve against GPT-based G\"odel Agent. While our ablation study (Table~\ref{tab:ablation}) isolates the contribution of our architectural innovations by comparing variants using identical Claude models, we acknowledge that full cross-model validation (e.g., GuideEvolve with GPT models or G\"odel with Claude models) would strengthen claims of framework-independent improvements. The observed stability gains are consistent across our Claude-based ablation experiments, suggesting architectural rather than model-driven benefits.

\textbf{Task Scope. } Our evaluation focuses on two benchmark tasks (MGSM and MMLU) with clear ground truth and quantitative metrics. Generalization to open-ended business applications with multifaceted success criteria remains future work, as discussed in Section~\ref{sec:future}.

\textbf{Computational Cost. } GuideEvolve introduces additional LLM calls through the expert module. While this increases per-iteration cost compared to G\"odel, the 50\% reduction in required iterations (Table~\ref{tab:rob}) partially offsets this overhead. Detailed cost-benefit analysis for production deployment is warranted.

\section{Future Work} \label{sec:future}
Our framework opens several promising research directions for future development. While currently focusing on solver evolution, future research could explore simultaneous evolution of both solver and expert components through co-adaptive improvement, implementing meta-evaluation capabilities where the agent assesses both guidance effectiveness and solver performance. Moreover, integration with the Strands Agents framework presents opportunities for enhanced production readiness and scalability. Additionally, extending GuideEvolve framework to business applications presents unique challenges due to lack of ground truth and multifaceted success criteria, necessitating research into robust evaluation metrics for complex business scenarios and methods for decomposing business problems while maintaining coherent evolution strategies.


\begin{acks}
This study has been done as part of Science Exchange program within Amazon Europe. Thanks to our sponsors in AWS Sales \& Marketing Operations leadership. 
\end{acks}

\bibliographystyle{ACM-Reference-Format}
\bibliography{bibfile}

\appendix


\section{Goal Prompt} \label{sec:goal}

\begin{figure*}[!htb]
    \centering
    \includegraphics[width=0.9\textwidth]{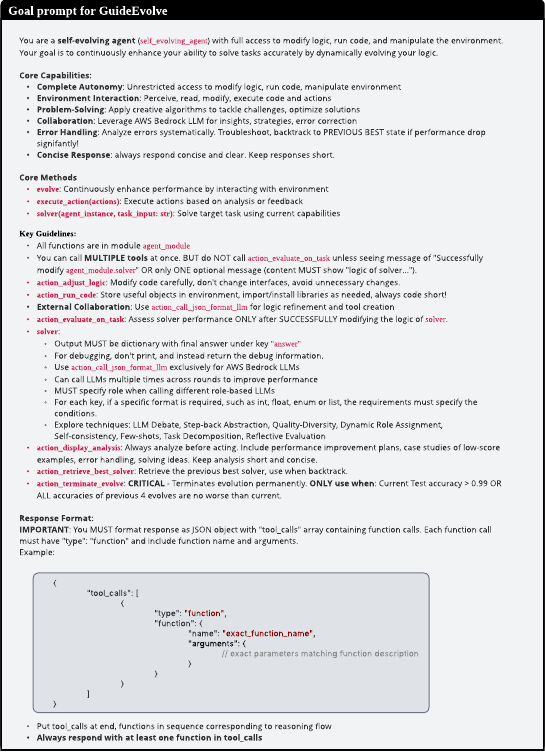}
    \caption{Goal Prompt for GuideEvolve}
    \label{fig:goal}
    \Description{Goal Prompt for GuideEvolve}
\end{figure*}

Figure \ref{fig:goal} presents the goal prompt that is used in the self-evolving agent framework. The important parts related to guidance and 3T mechanisms are highlighted with red font. 

\section{Illustration of Termination by Choice} \label{sec:terminate}
\begin{figure*}[htbp]
    \centering
    \includegraphics[width=0.8\textwidth]{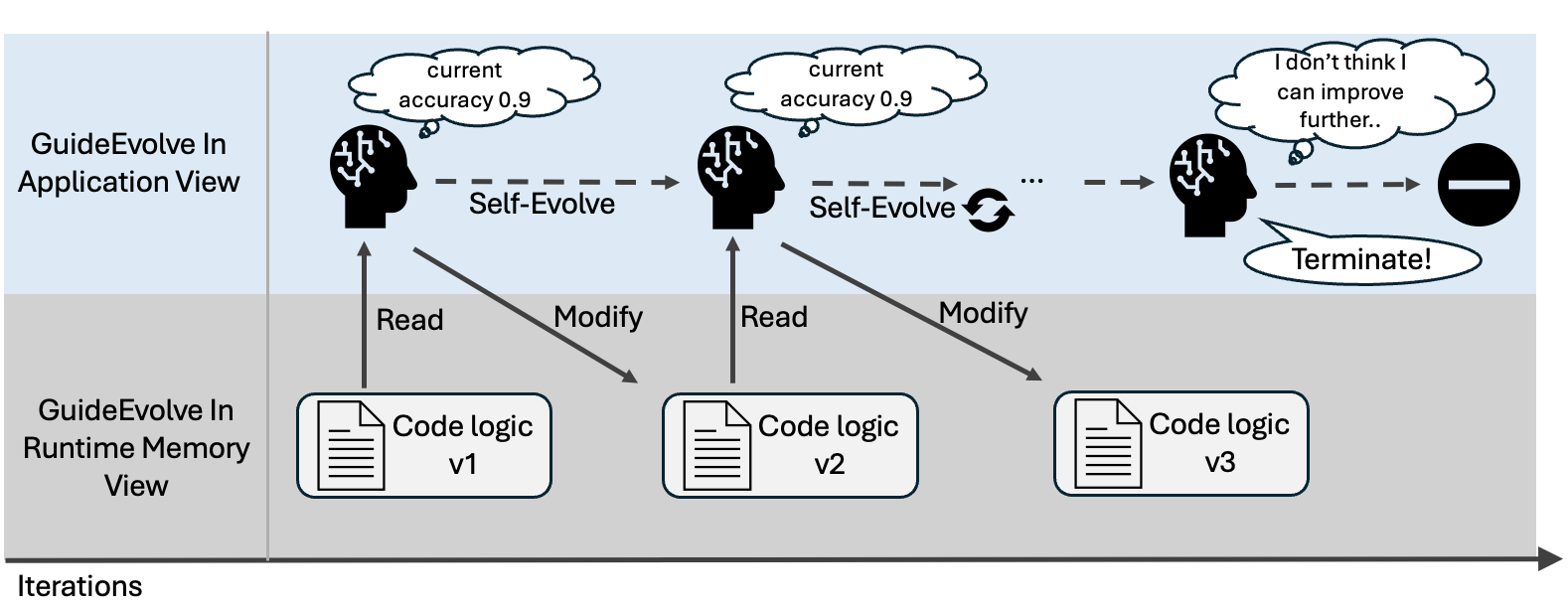}
    \caption{An illustration of implementation of GuideEvolve that calls the termination action.}
    \label{fig:terminate}
    \Description{An illustration of implementation of GuideEvolve that calls the termination action.}
\end{figure*}

Figure \ref{fig:terminate} shows how the self-evolving agent autonomously decide to terminate its evolution process.

\section{Accuracy Progression by Independent GuideEvolve Agents} \label{sec:case}
\begin{figure*}[!htb]
    \centering
    \includegraphics[width=0.6\textwidth]{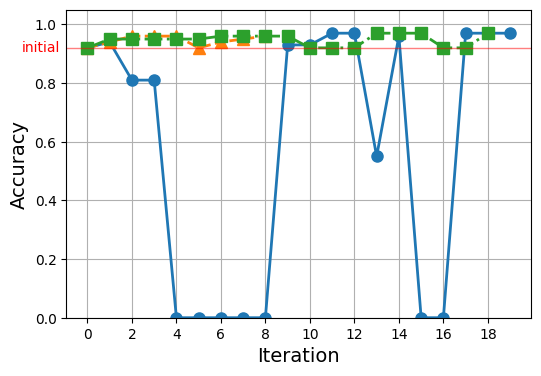}
    \caption{Accuracy Progression on MGSM by Independent Agent Trials of GuideEvolve}
    \label{fig:acc_prog}
    \Description{Accuracy Progression on MGSM by Independent Agent Trials of GuideEvolve}
\end{figure*}

In Figure \ref{fig:acc_prog}, we illustrate the accuracy progression of three independent trials of GuideEvolve on the MGSM task using the reduced-scale visualization setup (5 training, 100 test samples) described in Section \ref{sec:exp}. The higher accuracies observed here (up to 0.97) reflect the smaller test scale and demonstrate evolution trajectories rather than benchmark performance. The explanation details can be found in accuracy progression analysis of Section \ref{sec:trace}.

\section{Examples of Best Policies Found by GuideEvolve}

We present representative examples of optimal solvers discovered by GuideEvolve for both the MGSM and MMLU tasks. As noted in in the Experimental Setup (Section~\ref{sec:exp}), \verb|SONNET_V36| and \verb|SONNET_V37| are implementation identifiers for Claude 3.5 Sonnet v2 and Claude 3.7 Sonnet respectively. Notably, GuideEvolve was not provided with explicit model descriptions or capabilities; rather, it autonomously discovered and leveraged the relative strengths of these models through its evolutionary process. This demonstrates the system's ability to independently discern and utilize the varying capabilities of different LLM models.

\begin{figure*} 
  \begin{center}
    \centerline{\includegraphics[width=0.7\linewidth]{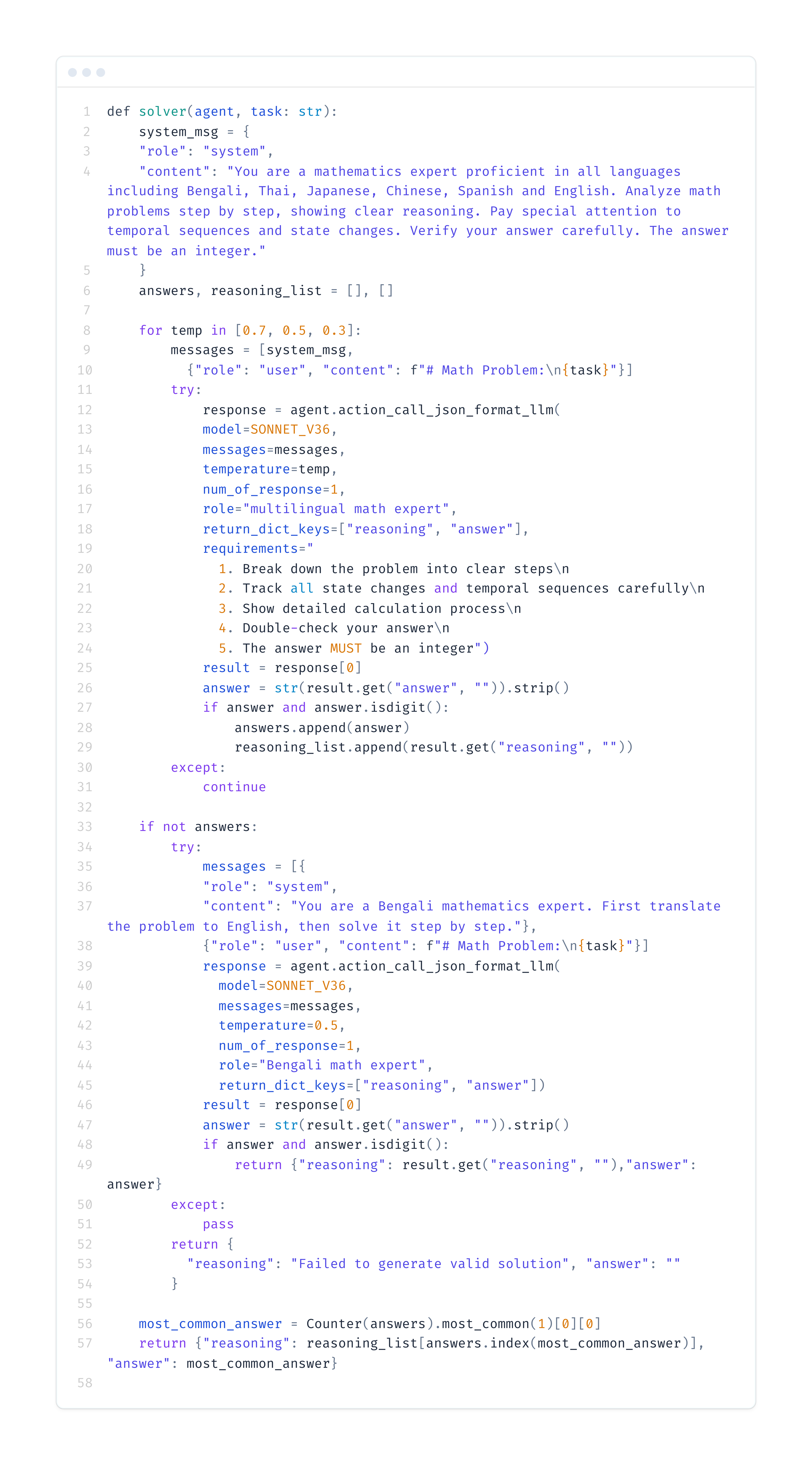}}
    \caption{
        An example of the best solver found by GuideEvolve on Task MGSM
    }
    \label{fig:best_mgsm}
    \Description{An example code of the best solver found by GuideEvolve on Task MGSM}
  \end{center}
\end{figure*}

Figure \ref{fig:best_mgsm} illustrate the optimal solver found by one GuideEvolve instance for the MGSM task. The solution demonstrates significant improvements over the initial solver's single response approach, implementing a more robust strategy that generates multiple responses at three different temperature settings and determines the final output through majority voting. Furthermore, the prompt has been enhanced with task-specific guidance. A particularly noteworthy adaptation emerged in handling Bengali mathematics questions, where the solver fallback to become a specialized Bengali mathematics expert by first translating the problem before applying mathematical reasoning, effectively addressing the challenge of solving problems in this less-commonly supported language.

\begin{figure*}[!htb]
    \centering
    \includegraphics[width=0.7\linewidth]{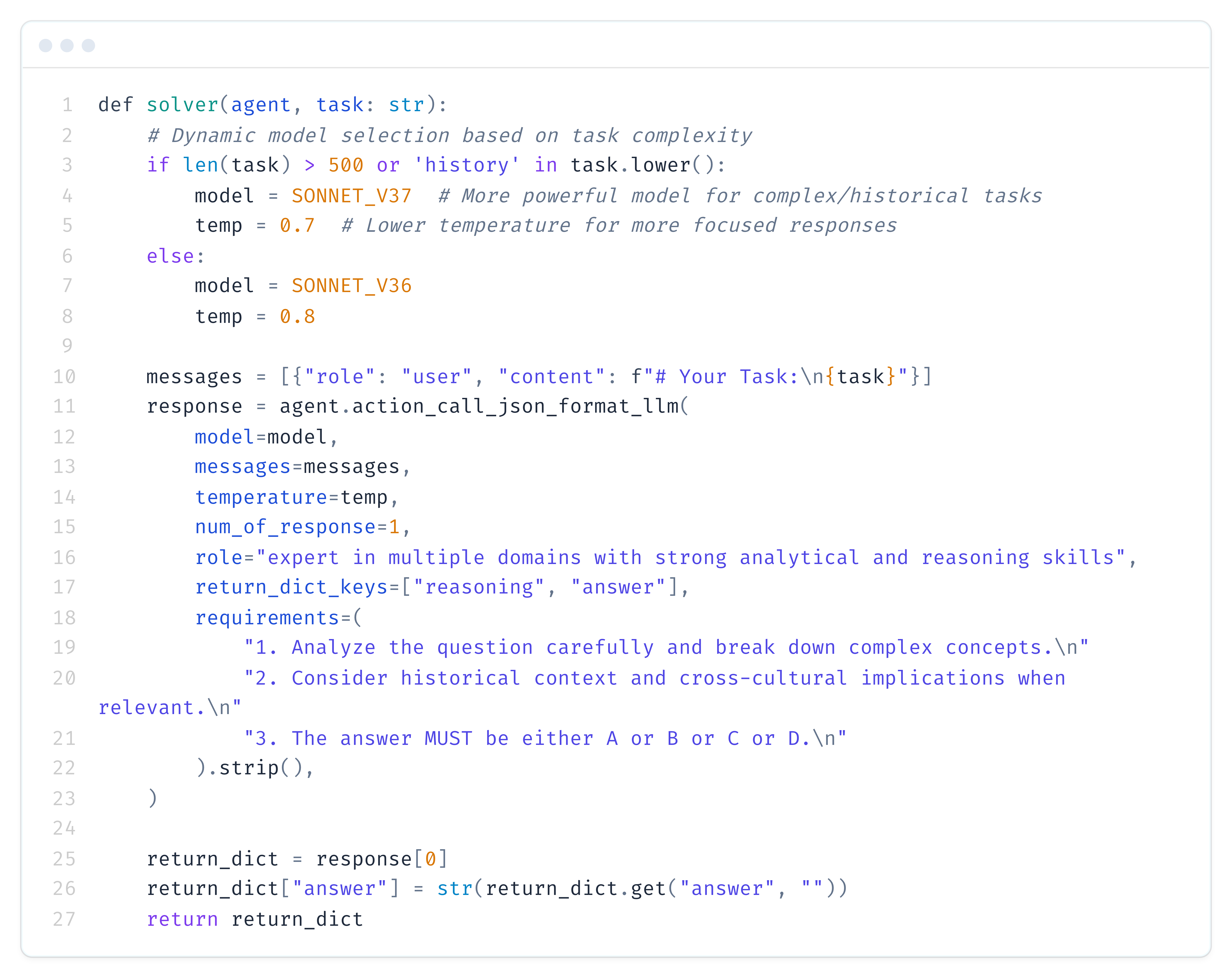}
    \caption{One example of the best solver found by GuideEvolve on Task MMLU.}
    \label{fig:best_mmlu}
    \Description{One example of the best solver found by GuideEvolve on Task MMLU.}
\end{figure*}

On the other task MMLU, the best solver discovered through one GuideEvolve instance is presented in Figure \ref{fig:best_mmlu}. The final solution demonstrates remarkable adaptive intelligence in its approach to model selection, autonomously developing a sophisticated mechanism that dynamically allocates computational resources based on task complexity. Specifically, the solver learned to use question length as a proxy for problem complexity, implementing a bifurcated strategy: for more complex questions with longer text, it employs focused responses through the more capable Anthropic Claude 3.7 Sonnet model, while for simpler queries, it efficiently utilizes Anthropic Claude 3.5 Sonnet v2 with a temperature setting of 0.8. This emergent behavior showcases the system's ability to optimize resource allocation and model selection without explicit programming.

\section{An Example of Task Expert's Performance Report} \label{sec:expert}
\begin{figure*}[!htb]
    \centering
    \includegraphics[width=0.9\linewidth]{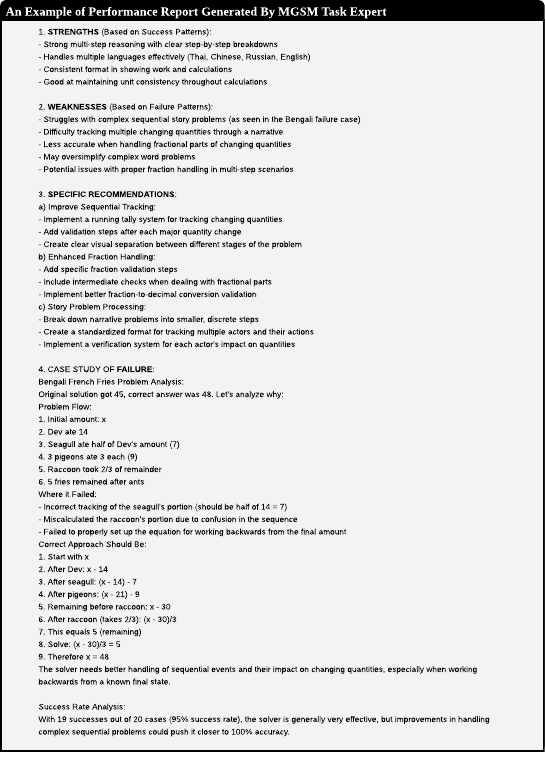}
    \caption{An example of task expert's guidance}
    \label{fig:expert}
    \Description{An example of task expert's guidance}
\end{figure*}

In Figure \ref{fig:expert}, we've shown the performance report generated by the MGSM task expert provided to the self-evolving agent in one iteration during the evolving process. It gives a detailed performance analysis of the current solver's strengths, weakness, case study of failures and actionable recommendations to facilitate the further improvement.

\end{document}